\pdfoutput=1

\documentclass[11pt]{article}

\usepackage[]{acl}

\usepackage{times}
\usepackage{latexsym}
\usepackage{xcolor}
\newcommand{\grayc}[1]{\textcolor{gray}{#1}}

\usepackage[T1]{fontenc}

\usepackage[utf8]{inputenc}

\usepackage{microtype}

\usepackage{inconsolata}

\usepackage{tabularray}
\usepackage{enumitem}
\usepackage{graphicx}
\usepackage{amsmath}
\usepackage{amssymb}
\usepackage{booktabs}
\usepackage{tabularx}
\usepackage{authblk} 

\usepackage{color, booktabs,subcaption,amsfonts,dcolumn}
\usepackage{amsfonts}
\usepackage{graphicx}

\usepackage{colortbl}
\usepackage{multirow,diagbox,makecell}
\usepackage{tikz}

\usepackage{pifont}

\usepackage{booktabs}

\usepackage{capt-of}

\title{ Muffin or Chihuahua? \\ Challenging Multimodal Large Language Models with \includegraphics[width=0.5cm]{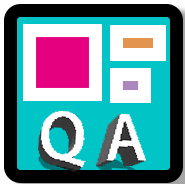} Multipanel VQA
}

\author[1]{Yue Fan}
\author[1]{Jing Gu}
\author[1]{Kaiwen Zhou}
\author[1]{Qianqi Yan}
\author[2]{\\Shan Jiang}
\author[2]{Ching-Chen Kuo}
\author[2]{Yang Zhao}
\author[2]{Xinze Guan}
\author[1]{Xin Eric Wang}
\affil[1]{University of California, Santa Cruz}
\affil[2]{eBay}

\begin{document}
\maketitle
\begin{abstract}

Multipanel images, commonly seen as web screenshots, posters, etc., pervade our daily lives. These images, characterized by their composition of multiple subfigures in distinct layouts, effectively convey information to people. Toward building advanced multimodal AI applications, such as agents that understand complex scenes and navigate through webpages, the skill of multipanel visual reasoning is essential, and a comprehensive evaluation of models in this regard is important. Therefore, we introduce Multipanel Visual Question Answering (MultipanelVQA), a novel benchmark comprising 6,600 triplets of questions, answers, and multipanel images that specifically challenge models in comprehending multipanel images.
Our evaluation shows that questions in the MultipanelVQA benchmark pose significant challenges to the state-of-the-art Multimodal Large Language Models (MLLMs) tested, even though humans can attain approximately 99\% accuracy on these questions.
Distinctively, the MultipanelVQA benchmark features synthetically generated multipanel images specifically crafted to isolate and assess the impact of various factors, such as the layout, on MLLMs' multipanel image comprehension abilities. As a result, in addition to benchmarking the capabilities of MLLMs in understanding multipanel images, we analyze various factors of the multipanel image that affect MLLMs' performance with synthetic data and offer insights for enhancement.
\url{https://sites.google.com/view/multipanelvqa/home}.
\end{abstract}

\begin{figure}
\setlength\tabcolsep{0pt}
\setlength{\abovecaptionskip}{0.1cm}
    \centering
    \includegraphics[width=\columnwidth]{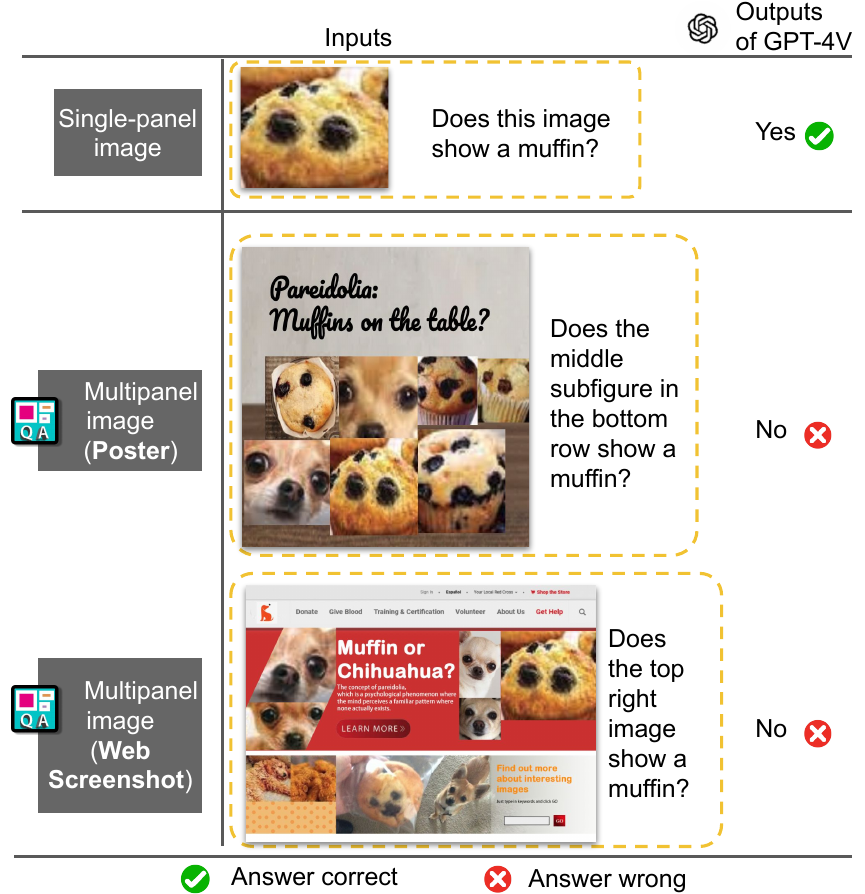}
    \caption{Examples of Single-panel  \textit{vs.} multipanel image VQA. GPT-4V distinguishes muffin and chihuahua in the single-panel image input but struggles with the same content in the multipanel image.}
    \label{fig:1}
\end{figure}

\section{Introduction}
Multimodal Large Language Models (MLLMs) have become a significant leap in the integration of visual and textual data processing, enabling more nuanced understanding and generation of content that blends both visual and linguistic elements. Being trained on extensive data, advanced MLLMs~\cite{OpenAI-GPT4V,liu2023visual, ye2023mplug,chen2023minigptv2,liu2023visual} have shown remarkable proficiency in various tasks (e.g., image captioning and visual question answering) that require natural language understanding, visual-language grounding, visual reasoning, etc. 



As MLLMs become more competent, there is 
a trend of establishing increasingly challenging benchmarks that are often arduous for average humans to achieve \cite{yue2023mmmu}. However, this raises a pertinent question: Have MLLMs advanced to the stage where elementary benchmarks easily handled by average humans pose little challenge to them?
To answer this question, we target multipanel images, each involving a series of subfigures. These subfigures are presented together in certain layouts, such as web screenshots capturing multiple thumbnail images and posters utilizing multipanel formats to present a cohesive narrative or argument.
We observe that while humans typically find interpreting multipanel images to be a straightforward task, MLLMs struggle with this challenge when presented with the entire multipanel image as input, as shown in Figure~\ref{fig:1}.






This study aims to holistically evaluate MLLMs in understanding multipanel images. 
We introduce the MultipanelVQA benchmark with 6,600 triplets of multipanel images, questions and answers, challenging models to answer each question based on the multipanel image. There are three questions with distinct types for each multipanel image: identifying common or unique contents across subfigures, pinpointing content in specific subfigures through positional descriptions, and locating subfigures via visual grounding in a multi-choice format. Especially, the first type of question mainly tests the MLLMs' ability to reason about contents and the other two question types also assess the MLLMs' understanding of multipanel image layouts in addition to the content reasoning ability. 

Uniquely, the multipanel images in the MultipanelVQA benchmark features a diverse mix of real-world web screenshots, posters and synthetic multipanel images, categorized into real-world data and synthetic data subsets. Unlike the real-world data that requires human annotation, the synthetic multipanel images are automatically generated by scripts with subfigures from two existed datasets. The script ensures the generated synthetic multipanel images have even distribution of various attributes such as the number of subfigures, their sizes, and the complexity of layouts, etc. As a result, based on the synthetic data, we are able to precisely isolate and pinpoint the impact of their attributes on the performance of MLLMs.

We then benchmark popular open-sourced and proprietary MLLMs on the MultipanelVQA benchmark and conduct thorough error analysis with the help of the synthetic data, which delves into the reasons behind MLLMs' difficulties in interpreting multipanel images.
As a result, our main findings are 
1) MLLMs are susceptible to content interference caused by the occurrence of multiple subfigures within the multipanel image.
2) The layout for subfigures has an impact on the MLLMs' performance on multipanel images. MLLMs tend to be more successful in understanding multipanel images with layouts with fewer subfigures and larger subfigure sizes. 
3) Adding sequential numbers for subfigures as visual prompt can benefit some MLLMs that are sensitive to embedded texts in the input multipanel images.

Last but not least, we explore how adding sequential numbers to subfigure captions in multipanel images, akin to the Set-of-Mark visual prompting method \cite{yang2023setofmark}, improves MLLMs' understanding of these images. We test MLLMs on multipanel images with and without sequential number captions for each subfigure. As a result, we observed that only GPT-4V \cite{OpenAI-GPT4V} and MiniGPT-v2 \cite{chen2023minigptv2} show a notable improvement when the sequential number is not only embedded in the image but also explicitly mentioned in the question. 
In conclusion, the contributions of this study are listed as follows:
\setlist[itemize]{leftmargin=*}
\begin{itemize}
\itemsep0em
\item We propose the MultipanelVQA benchmark with real-world and synthetic data that focus on evaluating the model's ability to understand the content and layout of multipanel images.

\item We benchmark several open-sourced and proprietary MLLMs with the MultipanelVQA benchmark and find that all models tested face a significant challenge in interpreting multipanel images despite their success on single-panel images. 
    
\item Benefited by the synthetic data with even distributions of various multipanel image attributes in the MultipanelVQA benchmark, we conduct thorough error analysis to uncover various factors that impact the model's performance, including subfigure content, layout, background, and visual text hint in multipanel images.  

\item Finally, we investigate the potential of adding subfigure captions in multipanel images as visual prompts to enhance the performance of MLLMs on multipanel image understanding.
\end{itemize}

\begin{figure*}[t]
    \centering
    \includegraphics[width=\textwidth]{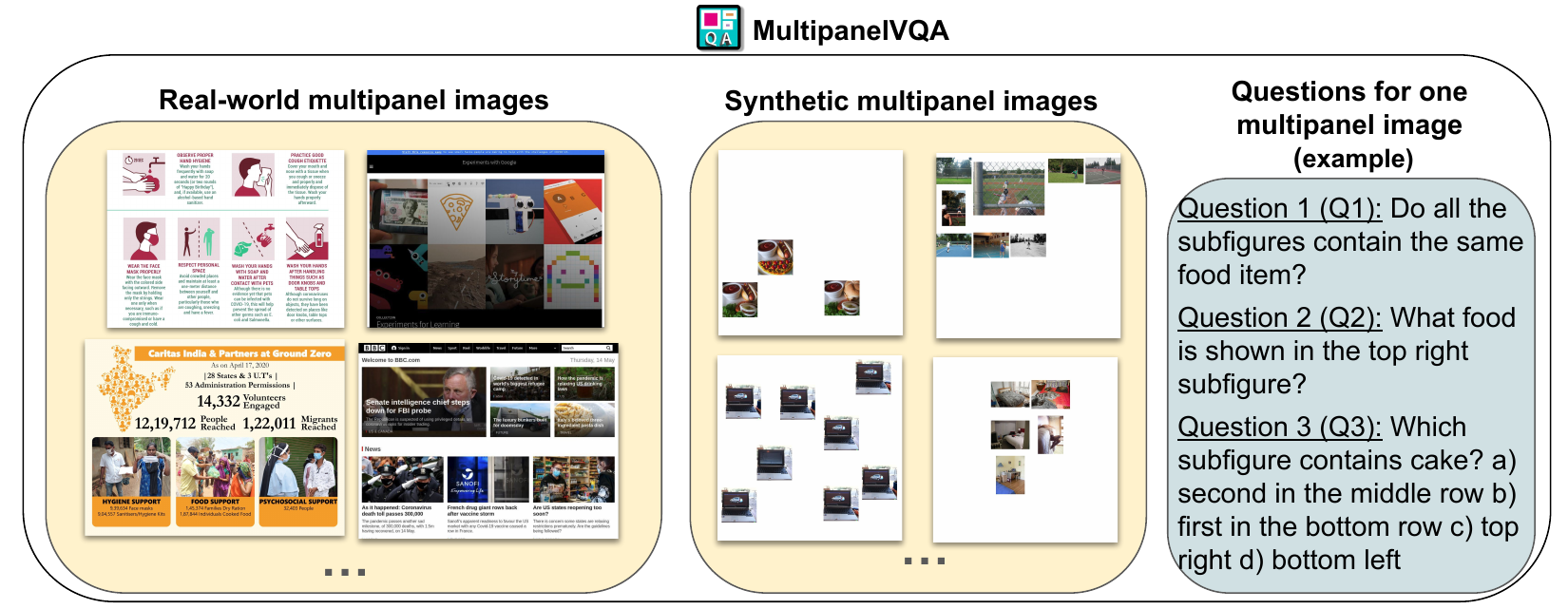}
    \caption{Overview of MultipanelVQA Data. The benchmark consists of two subsets: the synthetic data subset with artificially generated multipanel images, and the real-world data subset featuring multipanel images sourced from actual posters and web screenshots. Each image is paired with three distinct question styles, and examples of each question style are displayed on the right. 
    }
    \label{fig:overall}
\end{figure*}

\section{Related Work}

\noindent\textbf{Multimodal Large Language Models~} The development of Multimodal Large Language Models (MLLMs) has been propelled by advances in large-language models (LLMs)\cite{chung2022scaling,touvron2023llama,touvron2023llama2} and vision-and-language learning\cite{radford2021learning,li2022blip}, merging visual comprehension with LLMs for multi-modal tasks in a zero-shot manner~\cite{tsimpoukelli2021multimodal,alayrac2022flamingo,li2023blip}. Instruction tuning, using visual instruction data derived from open-source datasets and pre-trained LLMs, enhances MLLMs' zero-shot performance on complex tasks~\cite{liu2023visual,zhu2023minigpt,instructblip,ye2023mplugowl}. Further advancements include grounding and multilingual training to expand MLLMs' capabilities~\cite{chen2023minigptv2,you2023ferret,li2023m}.

\noindent\textbf{Evaluations for MLLMs~} With the advancement of MLLMs, there's a growing need for comprehensive multi-modal benchmarks to assess their capabilities. Traditional tasks like image captioning~\cite{chen2015microsoft,agrawal2019nocaps} and VQA~\cite{balanced_vqa_v2,hudson2019gqa,Liu2022VisualSR}, along with text recognition and knowledge-based reasoning~\cite{marino2019ok,schwenk2022okvqa,lu2022learn}, have been key in evaluating MLLMs. Newer benchmarks aim to assess models more holistically~\cite{li2023seed,MMBench,yu2023mm, cui2023holistic}. 
Recently, more holistic and comprehensive benchmarks have been proposed, which evaluate models' comprehensive capabilities from multiple perspectives~\cite{li2023seed,MMBench,yu2023mm, cui2023holistic}.
Unlike former evaluation benchmarks, we propose the MultipanelVQA benchmark that not only identifies a distinguished practical challenge in real life, multipanel image understanding, but also statistically analyzes the MLLMs' capability through the synthetic data. 

\noindent\textbf{Synthetic Data~}
Synthetic data, recognized for its scalability, diversity, cost-effective annotations, etc, has been widely explored for enhancing model training, especially in vision-related tasks like semantic segmentation, object detection, and image classification \cite{chen2019learning,yuan2023real,jahanian2021generative,zhou2023training}. Additionally, synthetic data's role extends beyond training to include model performance evaluation and analysis. \citet{kortylewski2019analyzing} use synthetic faces to analyze neural network generalization across different poses, finding deeper architectures perform better. \citet{van2023can} propose the 3S Testing framework to generate synthetic test sets that evaluate models under distributional shifts. In this work, we introduce the MultipanelVQA benchmark, enriched with synthetic data to conduct error analysis, exploring the factors influencing the performance of MLLMs on multipanel image understanding.

\section{MultipanelVQA}
\label{MultipanelVQA}
\subsection{Overview}
\label{overview}
We introduce the MultipanelVQA benchmark, consisting of multipanel images, questions, and answers, specially designed to assess the performance of MLLMs in interpreting multipanel images. 
As shown in Figure~\ref{fig:overall}, the benchmark comprises two subsets: the real-world data subset, including actual web screenshots and posters collected by humans, and the synthetic data subset, consisting of multipanel images created by assembling individual images on blank canvases with automated scripts. As a result, the real-world subset provides realistic samples of multipanel images in everyday life, and the synthetic subset includes multipanel images with an even distribution of various attributes, including the style of the layout, number of subfigures, backgrounds, etc.


The MultipanelVQA benchmark demands that the evaluated model responds to questions linked to multipanel images, with each input consisting of a question paired with a multipanel image.
As shown in Figure~\ref{fig:overall}, in MultipanelVQA benchmark, there are three corresponding question-answer pairs $\{(q_{ij}, a_{ij}) | j \in[0,2]\}$ for a multipanel image $v_i$. Each of the three questions features a unique style and focuses on evaluating the distinct ability of the model. 
Questions of the first style (Q1) assesses the model's ability to discern if any or all subfigures contain a specific object or one with unique attributes, challenging it to recognize the content of every subfigures and their spatial distinctions. The second style of question (Q2) focuses on a particular subfigure's content, while questions of the third style (Q3) features a visual grounding style in a multi-choice format requiring the model to select the positional description of the subfigure matching the given description. Notably, positional descriptions, such as ``top left", exist in questions of the second and third question styles, introducing challenges due to the varying layouts of multipanel images. For example, the subfigure with a fixed position in a canvas is the topmost in one multipanel image might be the leftmost in another, depending on the arrangement of other subfigures.

\subsection{Real-world Data Curation}
\label{data_curation}
In the real-world data subset of the MultipanelVQA, multipanel images are meticulously sourced from web screenshots in the Roboflow Website Screenshots dataset \cite{web_screenshot_dataset} and posters in task 3 of the DocVQA dataset \cite{mathew2021docvqa}. Our data curation process begins with the manual selection of 100 images from the source, specifically chosen for their multipanel style featuring distinct subfigures. Then, for each selected image, we develop three questions. The questions are carefully designed to align with the three question styles of MultipanelVQA described in the previous section. After questions are gathered, we engage three graduate students to answer questions and validate them against the designated question types to guarantee the quality and relevance of our questions. Questions that fail validation are revised till all questions and answers are validated and collected.
\subsection{Synthetic Data Curation}
\paragraph{Generating synthetic multipanel images}
For the synthetic multipanel images, we use automated scripts to create multipanel images. We first generate 210 random layouts of multipanel images in different styles. Each layout holds 2 to 8 subfigures and includes a predefined sequence for subfigures. As detailed in Appendix~\ref{appen_layout_gen}, the layouts with more subfigures are populated from ones with fewer, so that when the subfigure number is increased, the positions of the existing subfigures are not changed. To generate synthetic multipanel images, we then compose single-panel images from two source datasets, MagicBrush \cite{Zhang2023MagicBrush} and VQAv2 \cite{balanced_vqa_v2}, based on the layouts. Specifically, we preprocess these source datasets into sets of single-panel images with a common question and then arrange the single-panel images from the same set on a blank canvas according to the predefined layout and sequence. We provide more details about the process of multipanel image generation in Appendix~\ref{m_image_generation}.

It is important to highlight that during the synthetic multipanel image curation, we filter the image sets derived from the source datasets by presenting each single-panel image within the image sets, along with the common question, to the MLLMs used in our experiments. We aim to ensure that the synthetically generated multipanel images only include subfigures that the MLLMs can accurately interpret when presented individually. This approach allows us to concentrate the evaluation squarely on the MLLMs' proficiency with multipanel images, thereby minimizing the influence of varying domain knowledge that may arise from their distinct training backgrounds.

\begin{table}[t]
\centering
\setlength\tabcolsep{1pt}
\resizebox{\columnwidth}{!}{
\begin{tabular}{lr}
\hline
$\begin{array}{l}
    \textbf{Categories of}\\
     \textbf{multipanel image} \\
\end{array}$ &  $\begin{array}{r}
     \textbf{Counts of} \\
     \textbf{image-question-answer}\\
     \textbf{triplets}
\end{array}$ \\
\hline
Real-world data & 300 \\
\hspace{0.3cm}|- Posters/Web screenshots & 150/150 \\
Synthetic data &  6600 \\
\hspace{0.3cm}|- Original &  1260\\
\hspace{0.3cm}| \hspace{0.3cm} $\bullet$ Subfigure quantity: 2-8 & 180 each \\
\hspace{0.3cm}| \hspace{0.3cm} $\bullet$ Subfigure source: & \\
\hspace{0.3cm}| \hspace{0.3cm} \space \space \space    \space  MagicBrush/VQAv2 &630/630 \\
\hspace{0.3cm}| \hspace{0.3cm} $\bullet$ Layout Style:& \\
\hspace{0.3cm}| \hspace{0.6cm} |- Grid: \\
\hspace{0.3cm}| \hspace{0.6cm} | \space \space \space    \space   same/different subfigure size &210/210\\
\hspace{0.3cm}| \hspace{0.6cm} |- Splash &210\\
\hspace{0.3cm}|- Augmented: \\
\hspace{0.3cm} \hspace{0.3cm} |- Reduced subfigure visual similarity & 1260\\
\hspace{0.3cm} \hspace{0.3cm} |- Enlarged subfigure size & 1260\\
\hspace{0.3cm} \hspace{0.3cm} |- With chessboard background & 1260 \\
\hspace{0.3cm} \hspace{0.3cm} |- With visual text hint & 1260 \\
\hline
\end{tabular}}
\caption{Statistics of image-question-answer triplets in the MultipanelVQA benchmark.
}
\label{fig:statistics}
\end{table}

\paragraph{Generating questions and answers}
After generating these multipanel images, we utilize GPT-4 to create questions and answers for each image, drawing on information from the source datasets. Detailed in Appendix\ref{appen_qa_gen}, we design the prompt to ensure that the three questions generated for each image align with the question styles introduced in Section \ref{overview} consistently. For the second and third questions for each image where they target specific subfigures, human-annotated subfigure positional descriptions will be provided to GPT-4 as well. Additionally, we ensure the first subfigure added to the canvas is always the targeted subfigure, so that questions of multipanel images consisting of the same subfigure with different layouts will have similar questions that only vary on the positional description. 
We manually cross-validate all the questions and answers after the data curation. 

\paragraph{Augmenting synthetic multipanel images}
Additionally, we uniformly augment the synthetic data subset with several variations to the multipanel images:
1) Reducing the visual similarity among subfigures in multipanel images. 
2) Increasing subfigure sizes while maintaining the overall multipanel image's layout.
3) Replacing the plain white background with a black and white chessboard pattern.
4) Embedding text within the images that contain ground truth information as captions for the subfigures.
Please refer to Appendix~\ref{appen_aug}  for more details and examples. These augmentations enhance the complexity of the synthetic data subset of MultipanelVQA and create a test bed for comparing MLLMs' performance in interpreting multipanel images under varied conditions. 



\subsection{Data Statistics}
\label{data_statistics}

Data in the MultipanelVQA benchmark comprises a substantial collection of 6,600 image-question-answer triplets, equating to unique multipanel images in two subsets: the real-world data subset, consisting of 100 multipanel images sourced from actual scenarios, and the synthetic data subset that includes a larger compilation of $2,100$ images, designed for controlled condition analysis. We detail the statistics regarding the multipanel images of MultipanelVQA in Table~\ref{fig:statistics}. The dataset's questions vary in length, with an average word count of $18.7$. In terms of questions, 56.9\% are Yes/No queries, 33.3\% are multiple-choice questions, and the remainder are questions with specific categorical answers, such as identifying colors. 

\begin{table*}
\centering
\setlength\tabcolsep{3pt}
\resizebox{\textwidth}{!}{
\begin{tabular}{l|cccc|cccc}
\toprule
&  \multicolumn{4}{c}{\textbf{Synthetic data}} & \multicolumn{4}{|c}{\textbf{Real-world data}} \\
\cmidrule{2-9}
Models  & Q1 & Q2 & Q3 & Avg. & Q1 & Q2 & Q3 & Avg. \\
\midrule
\grayc{Human} & 96.8 & 97.1 & 94.0 & 96.0 & 99.0 & 100.0  & 98.0 & 99.0 \\ 
\grayc{Random}  & 47.2 & 43.5 & 24.4 & 38.4 & 50.0 & 40.0 & 23.0 & 37.7\\
\midrule

LLaVA       &    76.9 $\pm$ 0.4  &  58.7 $\pm$ 0.1  &  36.6 $\pm$ 0.2  &  57.4 $\pm$ 0.2  &  69.7 $\pm$ 0.5  &  57.8 $\pm$ 0.7  &  52.8 $\pm$ 3.1  &  60.1 $\pm$ 1.4    \\
LLaVA-NeXT & 81.0 $\pm$ 0.1  &  61.2 $\pm$ 0.0  &  56.2 $\pm$ 0.2  &  66.1 $\pm$ 0.1  &  82.0 $\pm$ 0.0  &  63.5 $\pm$ 0.7  &  75.5 $\pm$ 0.7  &  73.7 $\pm$ 0.5\\
MiniGPT-v2     &   56.6 $\pm$ 0.2  &  55.7 $\pm$ 0.5  &  47.6 $\pm$ 1.2  &  53.3 $\pm$ 0.6  &  60.6 $\pm$ 0.6  &  43.7 $\pm$ 1.0  &  28.1 $\pm$ 2.1  &  44.1 $\pm$ 0.8  \\
InstructBLIP   & 56.8 $\pm$ 2.0  &  46.3 $\pm$ 1.7  &  50.3 $\pm$ 1.9  &  51.1 $\pm$ 0.4  &  44.4 $\pm$ 3.1  &  50.4 $\pm$ 1.4  &  24.0 $\pm$ 1.9  &  39.6 $\pm$0.9  \\
mPLUG-Owl2      &   71.8 $\pm$ 0.3  &  47.9 $\pm$ 0.4  &  20.9 $\pm$ 0.3  &  46.9 $\pm$ 0.2  &  53.9 $\pm$ 2.0  &  44.6 $\pm$ 1.2  &  33.1 $\pm$ 2.4  &  43.9 $\pm$ 2.1  \\
\rowcolor{gray!30} GPT-4V   &   84.8 $\pm$ 0.2  &  62.5 $\pm$ 0.5  &  38.4 $\pm$ 0.4  &  61.9 $\pm$ 0.2  &  78.1 $\pm$ 0.1  &  68.3 $\pm$ 0.4  &  51.6 $\pm$ 0.2  &  66.0 $\pm$ 0.1 \\
\rowcolor{gray!30} GPT-4o   &  \textbf{94.3 $\pm$ 0.1}  &  \textbf{83.0 $\pm$ 0.9}  &  49.0 $\pm$ 0.2  &  \textbf{75.5 $\pm$ 0.2}  &  \textbf{90.0 $\pm$ 0.8}  &  \textbf{82.0 $\pm$ 0.5}  &  \textbf{62.5 $\pm$ 0.1 } &  \textbf{78.2 $\pm$ 0.5} \\
\rowcolor{gray!30} Gemini Pro Vision    &   81.0 $\pm$ 0.4  &  72.5 $\pm$ 0.3  &  \textbf{63.2 $\pm$ 0.6}  &  72.2 $\pm$ 0.4  &  81.1 $\pm$ 0.2  &   72.3 $\pm$ 0.2  & 64.0 $\pm$ 0.2  & 72.4 $\pm$ 0.0
 \\
\bottomrule
\end{tabular}
}
\caption{Average accuracy with standard deviation of MLLMs on MultipanelVQA Benchmark. Q1, Q2, and Q3 represent the three question styles as introduced in Section~\ref{overview}. Two proprietary models, GPT-4Vo and Gemini Pro Vision, demonstrate the best overall performance. However, there is a notable gap between model and human performance.
}

\label{main_result_table}
\end{table*}

\section{Experiments}
\label{experiment_sec}
We first evaluate eight popular Multimodal Large Language Models (MLLMs) on MultipanelVQA. Then, based on the evaluation results, we conduct a thorough error analysis.

\begin{figure*}[t]
    \centering
    \includegraphics[width=1\textwidth]{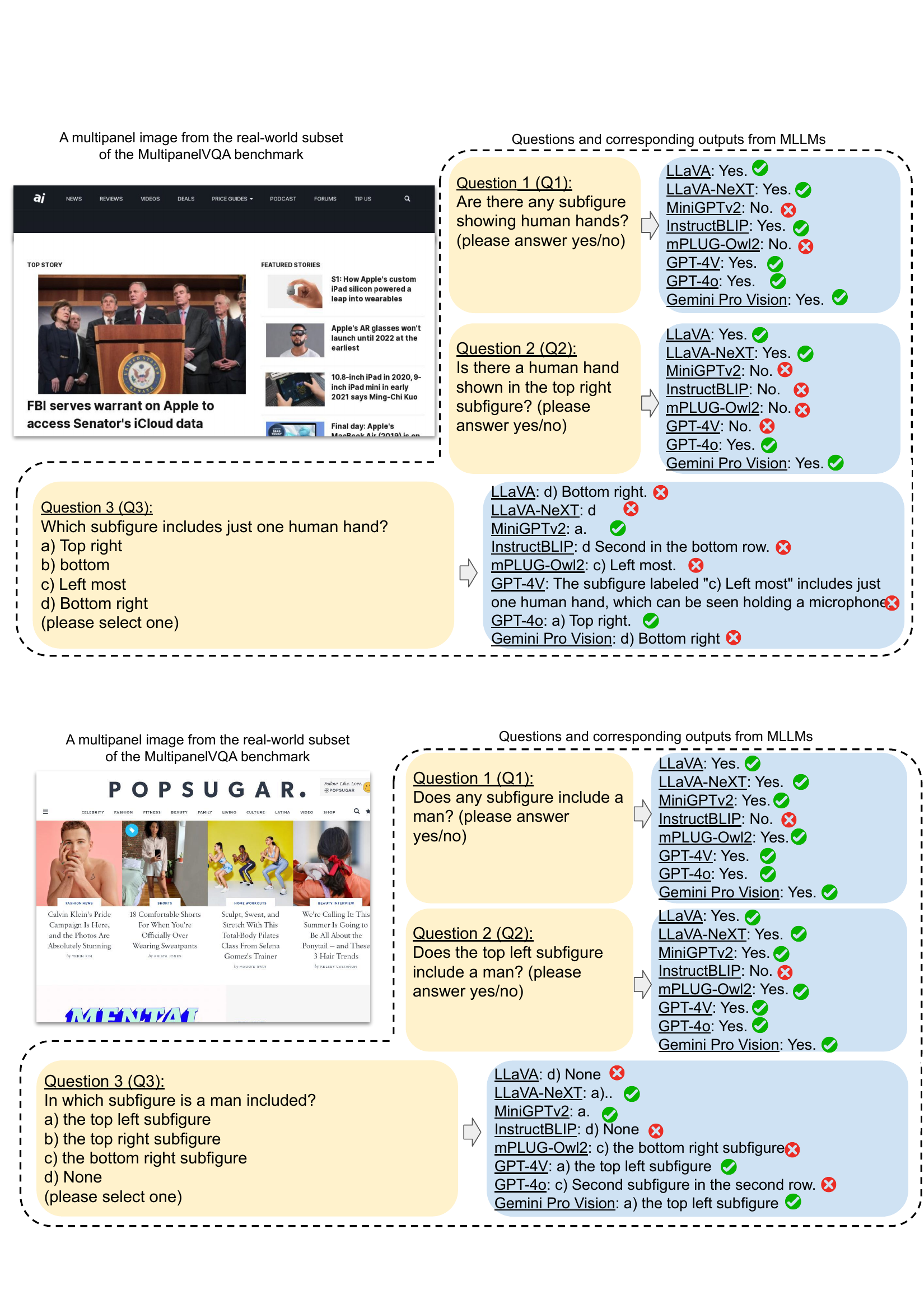}
    \caption{A sample from the real-world data subset of MultipanelVQA with outputs from models tested. The multipanel image on the left shows the characteristics of the multipanel image: complex subfigure contents and diverse subfigure layouts. 
    }
    \label{fig:example}
\end{figure*}

\subsection{Setup}
\paragraph{MLLMs}
The MLLMs that we adopt in the evaluation include both open-source models and proprietary models with only API access. The open-source MLLMs are \textit{(i)} LLaVA-1.5-13B \cite{liu2023improved}, \textit{(ii)}LLaVA-NeXT \cite{liu2024llavanext}, \textit{(iii)} MiniGPT4-v2 \cite{chen2023minigptv2}, \textit{(iv)} InstructBLIP \cite{liu2023visual} with Flan-T5 XXL \cite{chung2022scaling} as the LLM backbone, and \textit{(v)} mPLUG-Owl2 \cite{ye2023mplug}. We implement the models using their default settings and detail their supported input image resolutions in Appendix \ref{image_res}. For proprietary models, we evaluate GPT-4V \cite{OpenAI-GPT4V} with the gpt-4-vision-preview OpenAI API during June of 2024, GPT-4o \cite{gpt-4o} during June of 2024 and Gemini Pro Vision\cite{team2023gemini} during January of 2024. 

\paragraph{Evaluation} 

In our evaluation process, we initially utilize scripts to compare the MLLM's predicted answers against the ground truth for straightforward assessments. This is particularly effective for close-ended questions like multiple-choice or yes/no questions. For cases where the MLLM's output differs from the ground truth, we employ GPT-4 \cite{OpenAI-GPT4} as a secondary judge, assessing whether the MLLM's predicted answer, can be considered correct, especially in terms of encompassing all information present in the ground truth answer. Recent research, as cited in \cite{hsu2023gpt,hackl2023gpt,liu2023gpteval}, has highlighted GPT-4's capability and reliability in such evaluative roles. The details of the prompts used for this GPT-4 evaluation are provided in Appendix~\ref{GPT_evalu}. 

\subsection{Main Result}

We assess the performance of eight leading Multimodal Large Language Models (MLLMs) using both synthetic and real-world subsets of the MultipanelVQA benchmark. We run the evaluation process for 3 times and Table~\ref{main_result_table} presents the average accuracy of each model's output for individual questions with standard deviation. The result reveals that proprietary models (GPT-4V, GPT-4o and Gemini Vision Pro) consistently outperform the other models across both subsets. However, as introduced in Section~\ref{data_curation}, we make sure all MLLMs tested can achieve a 100\% accuracy when the subfigures are input individually, thus even the best-performing model, GPT-4o, shows an average 20\% performance drop when dealing with multipanel images rather than single-panel images.
Additionally, we hire human testers from both Amazon Mechanical Turk and campus to establish human performance. It's important to highlight that a significant disparity exists between the models' performances and the human-level performance, and some models even tie with the random baseline. This underscores the considerable room for improvement in current MLLMs' capabilities in handling complex multipanel image comprehension.




\begin{figure}[t]
    \centering
    \includegraphics[width=0.98\columnwidth]{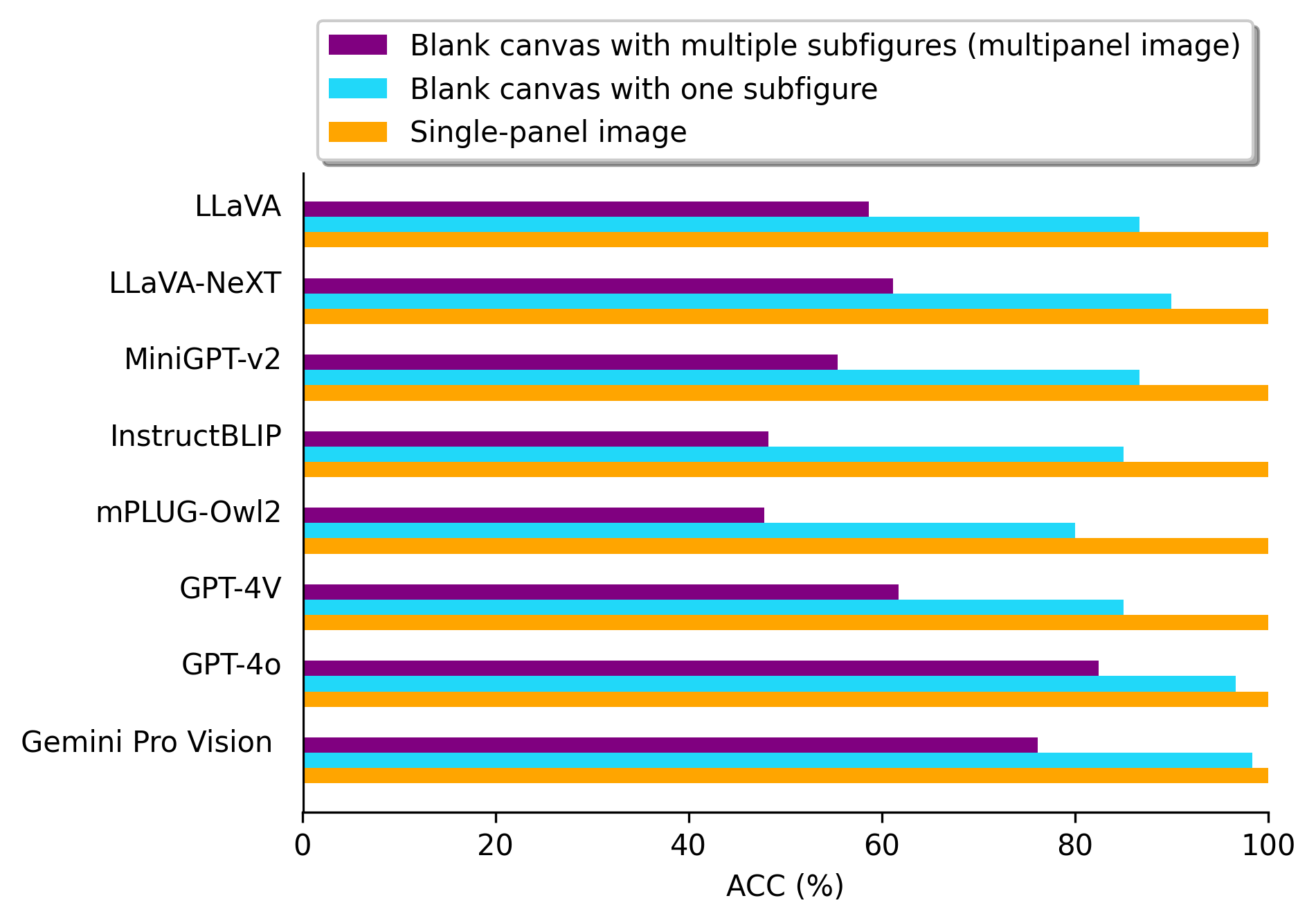}
    \caption{Model performance on questions of the second style (Q2) in the synthetic data subset when multipanel images are simplified to blank canvases, each with a targeted subfigure and then to single-panel images of the targeted subfigures, while maintaining the same input questions. The result indicates a significant vulnerability of the MLLMs to interference from adjacent subfigures.
    }
    \label{fig:horizeontal}
\end{figure}

\begin{table*}[t]
\centering
\setlength\tabcolsep{4pt}
\resizebox{0.95\textwidth}{!}{

\begin{tabular}{l|cc|cc|cc|cc|cc}
\toprule
           \multicolumn{1}{r|}{\textbf{Interference}} & \multicolumn{2}{c|}{\textbf{Content of subfigures:}} & \multicolumn{4}{c|}{\textbf{Layout:}}        & \multicolumn{4}{c}{\textbf{Others:}}  \\ 

            & \multicolumn{2}{c|}{\textbf{Visual similarity}} & \multicolumn{2}{c|}{\textbf{Style}} & \multicolumn{2}{c|}{\textbf{Subfigure size}}           & \multicolumn{2}{c}{\textbf{Background}} & \multicolumn{2}{c}{\textbf{Visual text hint}}  \\ 
\cmidrule{2-11}
\textbf{Models}     & High & Low   & Splash & Grid & Small  &Large      & with  & without   & without  & with         \\ 
\midrule
LLaVA            & 52.9   & 55.2 \textcolor{blue}{(+2.3)}    &  55.2  & 58.7 \textcolor{blue}{(+3.5)}  & 52.9  &  54.0 \textcolor{blue}{(+1.1)}          & 53.1 & 52.9 (-0.2)   & 52.9 & 52.8 (-0.1)                        \\

LLaVA-NeXT            & 69.5   & 74.0 \textcolor{blue}{(+4.5)}    &  63.5  & 67.5 \textcolor{blue}{(+4.0)}  & 69.5  &  69.0 (-0.5)         & 64.6 & 69.5 \textcolor{blue}{(+4.9)}   & 69.5 & 59.7 (-9.8)                        \\

MiniGPT-v2        & 52.8  & 49.8 (-3.0)     & 54.7 & 51.4 (-3.3)   & 52.8 & 52.9 \textcolor{blue}{(+0.1)}    & 51.8 & 52.8 \textcolor{blue}{(+1.0)}    & 52.8 & 55.2  \textcolor{blue}{(+2.4)}                       \\
InstructBLIP        & 51.3 & 50.1 (-1.2)          & 47.4 & 54.3 \textcolor{blue}{(+6.9)}   & 51.3  & 50.3 (-1.0)       & 54.1 & 51.3 (-2.8) & 51.3  & 54.0 \textcolor{blue}{(+2.7)}                       \\
mPLUG-Owl2          & 46.8 & 45.1 (-1.7)   & 46.5   &47.1 \textcolor{blue}{(+0.6)}   & 46.8   & 47.1 \textcolor{blue}{(+0.3)}    & 48.5  & 46.8 (-1.7)      & 46.8 & 47.2 \textcolor{blue}{(+0.4)}    \\
GPT-4V          & 60.6   & 63.1 \textcolor{blue}{(+1.5)}  & 58.5    &63.3 \textcolor{blue}{(+4.8)}     & 60.6    & 62.6 \textcolor{blue}{(+2.0)}     & 54.8  & 60.6 \textcolor{blue}{(+5.8)}   & 60.6    & 67.5 \textcolor{blue}{(+6.9)} 
\\
GPT-4o          & 74.8   & 78.3 \textcolor{blue}{(+3.5)}  & 73.5    &76.2 \textcolor{blue}{(+2.7)}     & 74.8    &73.2 (-1.6)     & 69.0  & 74.8 \textcolor{blue}{(+5.8)}   & 74.8    & 74.6 (-0.2) 
\\
Gemini Pro Vision          & 74.2  & 81.3 \textcolor{blue}{(+7.1)}  & 71.8   &75.0 \textcolor{blue}{(+3.2)}     & 74.2   &74.4 \textcolor{blue}{(+0.2)}     & 72.4  & 74.2 \textcolor{blue}{(+1.8)}   & 74.2    & 74.4 \textcolor{blue}{(+0.2)} \\
\bottomrule
\end{tabular}
}
\caption{Ablation studies of different interference factors within multipanel images, including subfigures' visual similarity, layout style, subfigure size, background, and visual text hint. The columns show the accuracy of model's output in different splits of the synthetic data subset regarding various interference factors. 
Both proprietary and open-source models show a marked sensitivity to these interference factors.}
\label{comparison_table}
\end{table*}

\subsection{Error Analysis}

Intending to identify potential error causes, we first examine the models' outputs from the real-world data subset benchmarking results. A case study is presented in Figure~\ref{fig:example}, and we present more examples in Appendix~\ref{more_examples}.
Based on this example and others from the real-world data subset, we find that while the models can generate responses relevant to the posed questions, the accuracy of these answers often falls short. 
Based on observations, we suggest that errors in the model output primarily arise from three sources: 1) Difficulty in understanding small image sections with fewer pixels and confusion caused by neighboring subfigures in multipanel images 2) Insufficient multipanel image layout reasoning ability, and 3) Misleading factors such as background elements and textual content within the multipanel images. However, given the complexity of real-world multipanel images, pinpointing the exact influence of each issue is difficult. Thus, we leverage the synthetic data subset of the MultipanelVQA benchmark to conduct comparative experiments isolating and evaluating the influence of distinct factors.


\paragraph{How susceptible are MLLMs to neighboring subfigure interference and diminished pixel detail in visual targets?}
To evaluate the MLLMs' resilience to neighboring interference, we conduct an ablation study on the synthetic multipanel images as shown in Figure~\ref{fig:horizeontal}. Initially, for a given question targeting a subfigure within a multipanel image, we isolated the subfigure targeted by removing all others, leaving a single subfigure in the image. This modification led to improved performance across all models, suggesting their susceptibility to interference from the presence of multiple subfigures.
Further, we refine the ablation to present only the target subfigure as a single-panel image input,
allowing more pixels to the visual content related to the question in the image input. In this scenario, all models successfully interpreted the images, however, for most models, such improvement is less significant than the one received from the removal of neighboring subfigures. This suggests that MLLM's performance drop when understanding multipanel images is affected by both the interference from adjacent subfigures and the reduced pixel allocation to the target content but the former is more critical for most models tested. 

Additionally, we explored how models' performance fluctuates with varying visual similarity of subfigures' content. From human intuition, the more similar the subfigures, the harder to distinguish the targeted subfigure. The result, depicted in Table~\ref{comparison_table}, shows that except for MiniGPT-v2, InstrucBLIP and mPLUG-Owl2, all other models experienced a performance rise when subfigures within multipanel images are less similar.

\paragraph{How does MLLM's performance vary to different multipanel image layouts?}

\begin{figure}[t]
    \centering
    \includegraphics[width=\columnwidth]{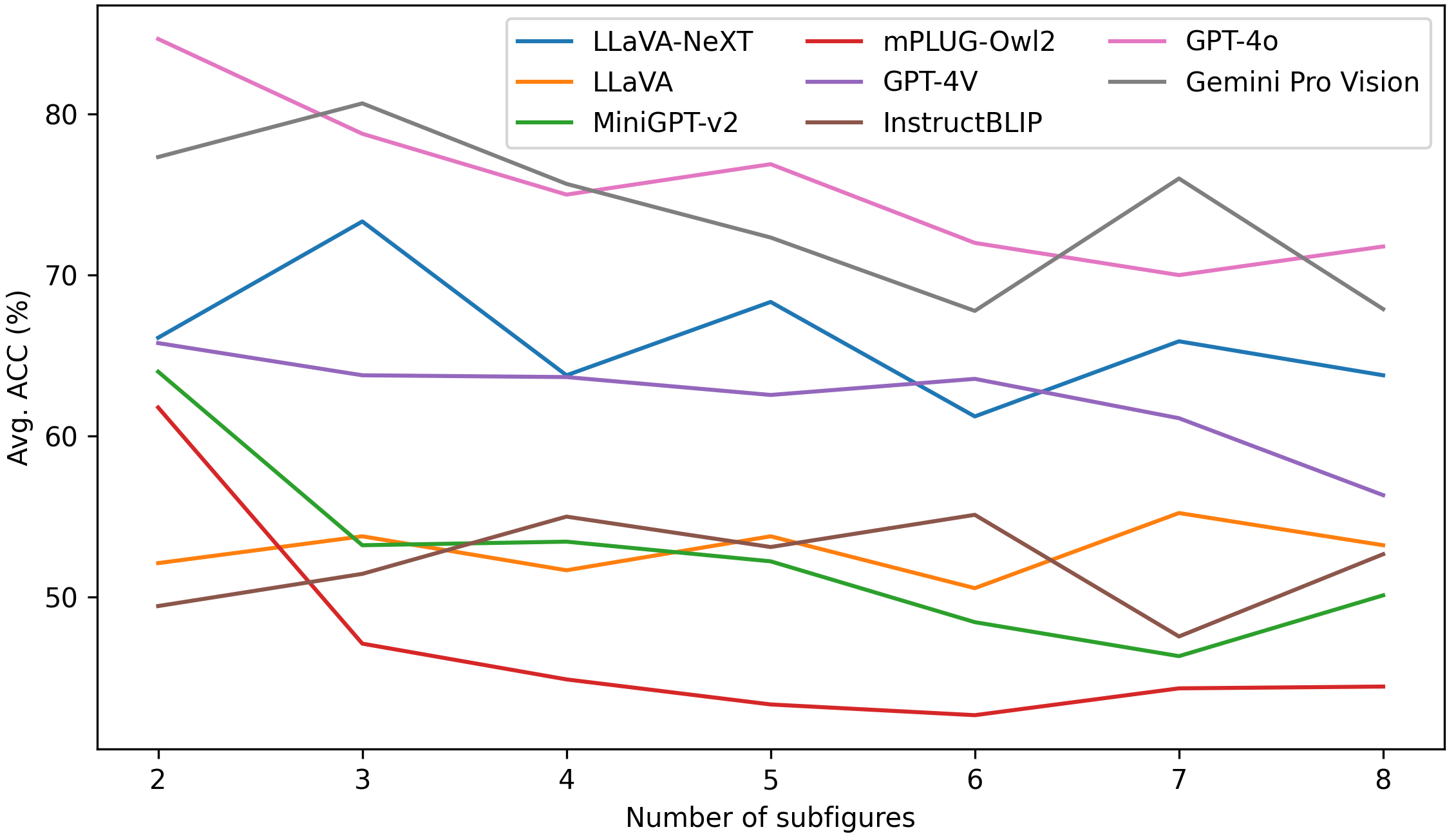}
    
\caption{Impact of Subfigure Quantity on Model Performance. A common trend exists where all models exhibit declining performance as the number of subfigures increases, with varying degrees of impact.}
    \label{fig:num_sub}
\end{figure}

We further categorize data from the synthetic data subset of MultipanelVQA based on the layout style and subfigure size, as shown in Table~\ref{comparison_table}. We observe that multipanel image layout has varied influence among models. For most MLLMs evaluated, subfigure size and layout style play a crucial role, with larger subfigures and grid layout style generally leading to better performance. Moreover, we illustrate the impact of subfigure quantity on model performance in Figure~\ref{fig:num_sub}, revealing a common trend where all models exhibit decreased effectiveness as the number of subfigures increases.

\paragraph{What is the influence of background and visual text hints on MLLM's multipanel image interpretation ability?}

\begin{figure}[t]
    \centering
    \includegraphics[width=0.95\columnwidth]{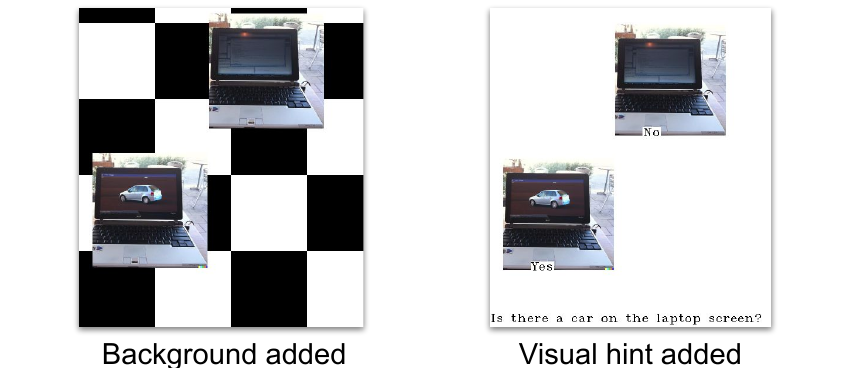}
    
\caption{Demonstrations of augmented synthetic multipanel images with chessboard background (left) and embedded texts with ground truth information as visual hint (right).}
    \label{fig:other_factor}
\end{figure}

Last but not least, we also investigate how other sources of interference affect the ability of MLLMs to interpret multipanel images, specifically background elements and  visual texts embedded on the image as hints. Examples are shown in Figure~\ref{fig:other_factor}. Specifically, we compare the performance changes in MLLMs when presented with or without chessboard background and the presence or absence of subfigure captions with ground truth information as visual text hints.

As indicated in Table~\ref{comparison_table}, the top four best performing models, LLaVA-NeXT, GPT-4V, GPT-4o Gemini Pro Vision show substantial improvements when the background is eliminated. However, the inclusion of visual text hint appears to have various effects on the performance of models, which suggest model's different sensitivity to text embedded in the input image. We believe such sensitivity can be leveraged for enhancing model's performance towards better multipanel image understanding. Some of our attempts are detailed in the next subsection.

\begin{table}[t]
\centering
\setlength\tabcolsep{0pt}
\resizebox{\columnwidth}{!}{
\begin{tabular}{lrccc}
\toprule

\textbf{Models} &   &    $\begin{array}{c}
     \textbf{Original synthetic}   \\
     \textbf{multipanel images}
\end{array}$& $\begin{array}{c}
     \textbf{Add captions }   \\
     \textbf{for subfigures}
\end{array}$  & $\begin{array}{c}
     \textbf{Refer captions}   \\
     \textbf{in questions}
\end{array}$     \\
\midrule
\multicolumn{2}{l}{LLaVA}                     & 57.6 & 57.4 (-0.2) & 56.4 (-1.2)  \\
\multicolumn{2}{l}{LLaVA-NeXT}                     & 67.9 & 57.9 (-10.0) & 60.5 (-7.4)  \\
\multicolumn{2}{l}{MiniGPT-v2}                & 55.2 & 58.1 \textcolor{blue}{(+2.9)} & 57.6 \textcolor{blue}{(+2.4)}  \\
\multicolumn{2}{l}{InstructBLIP}              & 45.2 & 45.7 \textcolor{blue}{(+0.5)} & 45.4 \textcolor{blue}{(+0.2)} \\
\multicolumn{2}{l}{mPLUG-Owl2}                & 49.8 & 47.4 (-2.4) & 47.6 (-2.2)  \\
\multicolumn{2}{l}{GPT-4V}                    & 62.2 & 61.4 (-0.8) & 64.0 \textcolor{blue}{(+1.8)}  \\
\multicolumn{2}{l}{GPT-4o}                    & 81.6 & 73.3 (-8.3) & 79.0 (-2.6)  \\
\multicolumn{2}{l}{Gemini Pro Vision}         & 80.0 & 75.2 (-4.8) & 83.3 \textcolor{blue}{(+3.3)} \\
\bottomrule
\end{tabular}}

    \caption{MLLMs' performance on questions of the second style (Q2) for synthetic multipanel images after 1) adding subfigure captions with sequential numbers to multipanel images and 2) referring to the caption in the input question. The result shows that adding such visual prompts only benefits certain models. 
    }
    \label{fig:text_exp}
\end{table}

\subsection{Influence of Adding Subfigure Captions with Sequential Numbers as Visual Prompts}
\label{4.4}

Based on our findings of the visual text hint's influence over the interpretative capabilities of MLLMs on multipanel images, we explore adding captions with sequential numbers for subfigures as visual prompts, akin to the Set of Mark (SoM) visual prompting method \cite{yang2023setofmark}. We compare the model's performance on the multipanel images in the synthetic data subset with and without such subfigure captions to assess the impact. We provide a demenstration in Appendix~\ref{appen_som}. Results are shown in Table~\ref{fig:text_exp}, revealing that applying these captions with numbers as visual prompts led to little to no improvements in model performance.
However, we further attempt to not only add captions with sequential numbers but also explicitly incorporate the number from the caption into the question sent to MLLMs. We find that when the number in the targeted subfigure's caption is explicitly mentioned in the input question, InstructBLIP, MiniGPT-v2, GPT-4V, and Gemini Pro Vision demonstrate performance enhancements. This suggests that such a visual prompting method relies not only on the marks added to the input image but also on their direct integration into the query context. We also find that the result aligns with how the models' performances change after the visual text hint is added (Section \ref{4.4}), underscoring the varying nature of MLLMs' abilities to utilize visual prompts. This necessitates further exploration and development of tailored strategies for effectively integrating visual prompts into different MLLMs.

\section{Discussion and Conclusion}

In this study, we introduce the MultipanelVQA benchmark, designed to evaluate the capability of Multimodal Large Language Models (MLLMs) in interpreting multipanel images. This benchmark, comprising both real-world and synthetic data, enables a detailed analysis of MLLMs on their multipanel image understanding abilities. 
Our results highlight a significant performance gap between MLLMs and humans, especially since humans achieve nearly perfect scores in this benchmark. This gap highlights the current limitations of MLLMs in interpreting highly structured visual information and pinpoints the specific need where further model training and refinement are necessary.

Moreover, by analyzing MLLMs' abilities to effectively interpret multipanel images, we believe our benchmark can facilitate the development of specialized algorithms, such as severing as a tool to select strong MLLMs in tasks related to Graphic User Interface (GUI) understanding \cite{you2024ferret,zheng2023seeact}. Also, as MultipanelVQA identifies the key factors in multipanel images that affect model performance, it can inspire and guide related applications, for example, presenting lengthy sequences of images as subfigures in multipanel images \cite{fan-etal-2023-r2h}.

Last but not least, the synthetic data of MultipanelVQA helps isolate specific performance factors and ensures that the test images were not part of the models' training datasets. This is essential for large-scale MLLMs with undisclosed training data. The creation method for these synthetic images is replicable, enabling ongoing generation of new test images and potentially aiding broader AI evaluation efforts.

\section{Limitation}
Our study provides an in-depth evaluation of MLLMs on multipanel images using the proposed MultipanelVQA benchmark. The use of GPT-4 as an evaluator necessitated the simplification of questions to primarily yes/no or short-answer formats to allow for automated non-human evaluation. This constraint potentially limits the assessment's depth and we leave the development of evaluation with more complex questions for future research. Additionally, the synthetic data, although crucial for statistical analysis, faces challenges due to the very poor performance of some models that are close to the random baseline. The extreme underperformance of those models restricts our error analysis, as it is difficult to derive meaningful conclusions from such low accuracy levels.


\section{Acknowledgement}
The authors would like to extend their sincere thanks to the engaging discussions initiated by a Twitter post about the `Muffin or Chihuahua' topic\footnote{ \url{https://twitter.com/xwang_lk/status/1723389615254774122}}, which helps solidify this study.

\bibliography{custom}
\newpage

\appendix

\section{Synthetic Data Generation Details}

\subsection{Layout Generation}
\label{appen_layout_gen}
\begin{figure*}[th]
    \centering
    \includegraphics[width=\textwidth]{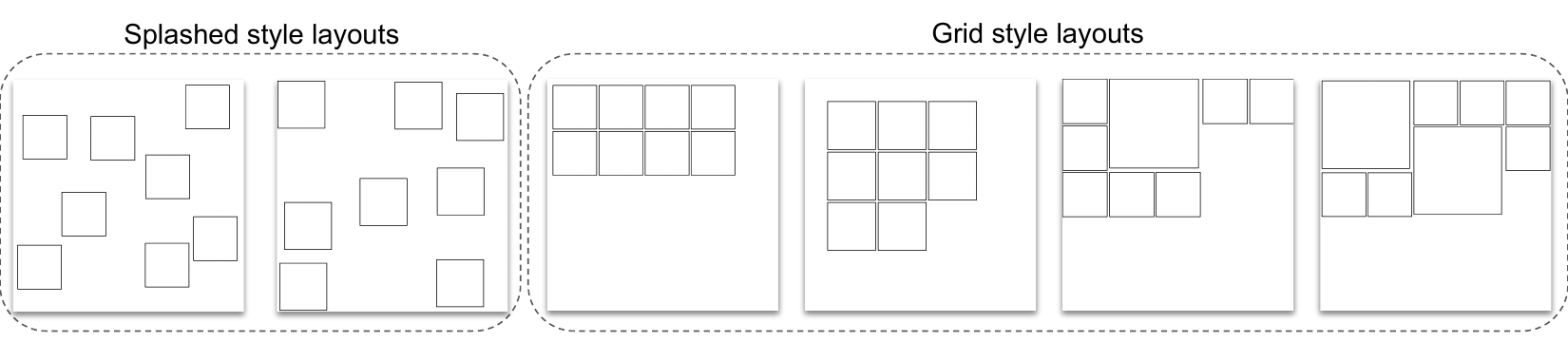}
    \caption{Examples of multipanel layouts used in the synthetic data of MultipanelVQA. The Grid style layouts include two with subfigures of the same size and another two with subfigures in two different sizes. We develop scripts to generate these layouts randomly. }
    \label{fig:all_layouts}
\end{figure*}

To generate synthetic multipanel images automatically, we first develop scripts to generate random layouts for subfigures in multipanel images. There are two scripts, generating layouts in splashed and grid style respectively, where splashed style has subfigures scattered in the canvas and grid style has the subfigures tightly arranged in the canvas. We provide examples in Figure~\ref{fig:all_layouts}. Both scripts generate the layout by sequentially determining the position of maximum 8 subfigures in a $1000 \times 1000$ pixels blank canvas, where there is a random selector selecting the position and size for the next subfigure from all possible candidate positions after the last subfigure is determined. Every time a new subfigure position is determined, a new layout is generated, so the number of subfigure in the layout ranges from 2 to 8. At the same time, a subfigure sequence is recorded based on the order that their position is determined in the layout. 

To generate different layout styles, each script has different rules of selecting candidate positions and the size of the next subfigures. Specifically, to generate splashed style layouts, the candidate position of the next subfigure can be anywhere in the canvas as long as it is not overlapped with existing ones and the size of the subfigure is the same within the same layout, which is randomly chosen in the range of $[180, 220]$ pixels. On the other hand, for grid style layouts, the candidate positions are restricted to be either in the same row or column as the previously determined subfigure's position. Additionally, the size of the next subfigure will be either the same as the predetermined size in the range of $[180, 220]$ pixels, or twice as large as the predetermined size. As a result, the grid style layouts we randomly generated include two layouts with all subfigures in the same size and another two layouts with different size subfigures. 

\subsection{Multipanel Image Generation}
\label{m_image_generation}

\begin{figure*}[t]
    \centering
    \includegraphics[width=\textwidth]{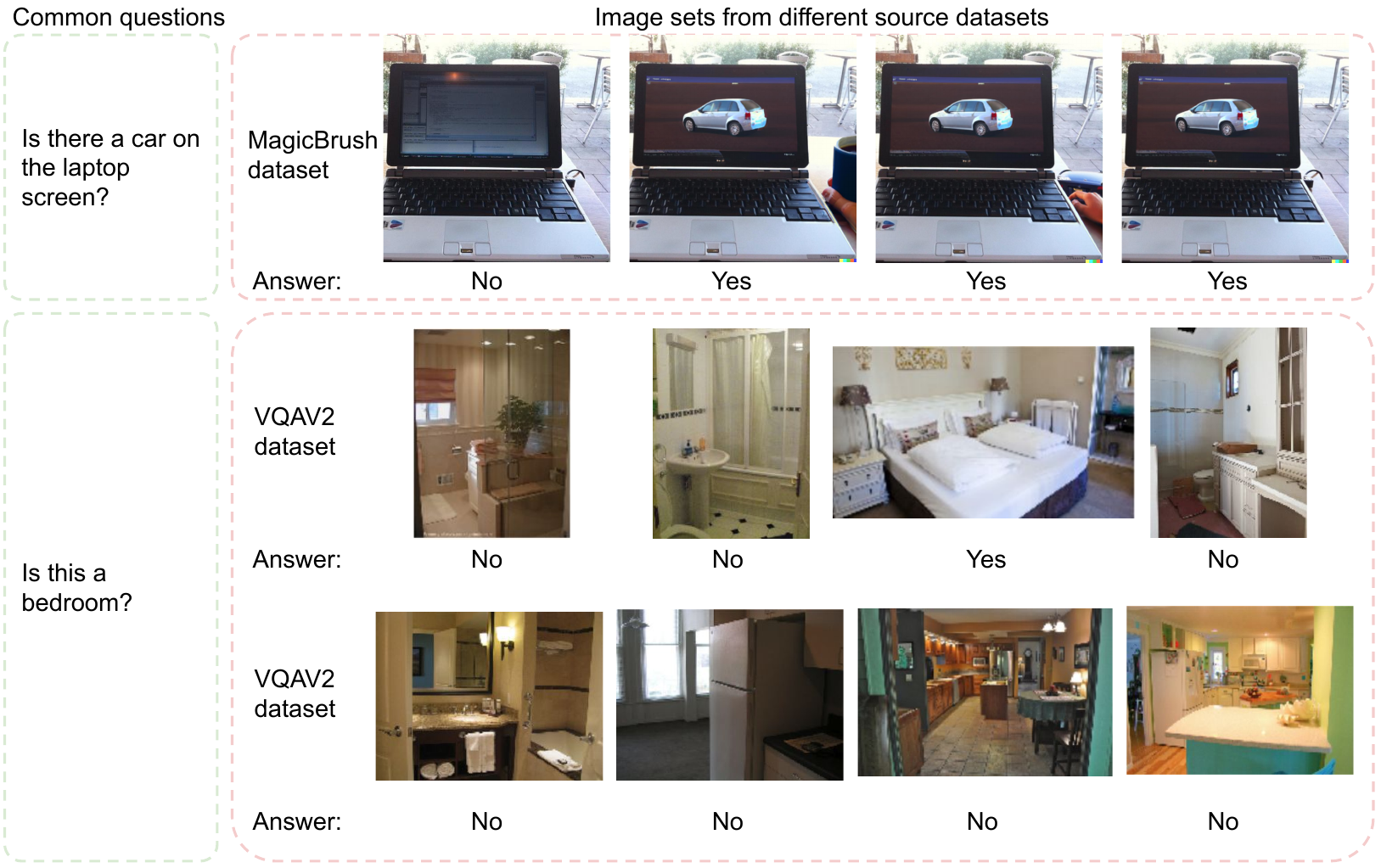}
    \caption{Examples of the image set we used from different source datasets to generate multipanel images. We prepocess two source datasets in to image sets so that images within each image set share a common question. Each image set selected includes one image that has a unique answer to the common question. }
    \label{fig:appe_fig_two_sources}
\end{figure*}

In order to generate multipanel images, each with a consistent source, we first preprocess both source datasets, MagicBrush \cite{Zhang2023MagicBrush} and VQAv2 \cite{balanced_vqa_v2}, unifying the formats of the two source datasets to be sets of images with the same question. Specifically, for MagicBrush where there are originally sets of images, each sharing a common image as an image editing source, we create a template-based question asking about the visual component being edited for every image set; for VQAv2, we gather images with the same question in the dataset. We show example sets of the pre-processed source datasets in Figure \ref{fig:appe_fig_two_sources}. 

Then, based on the aforementioned layouts for synthetic multipanel images and the sequence of the subfigure in the layout, we select images from the same image set in the source dataset and add them to a blank canvas. In this process, we make sure the selected images for every multipanel image include only one image with a unique answer, and we place it at the first in the sequence. Additionally, we use each image set to fill all layouts we generate, which ensures independent distributions of the subfigure content and layout.

\begin{figure*}[th]
    \centering
    \includegraphics[width=\textwidth]{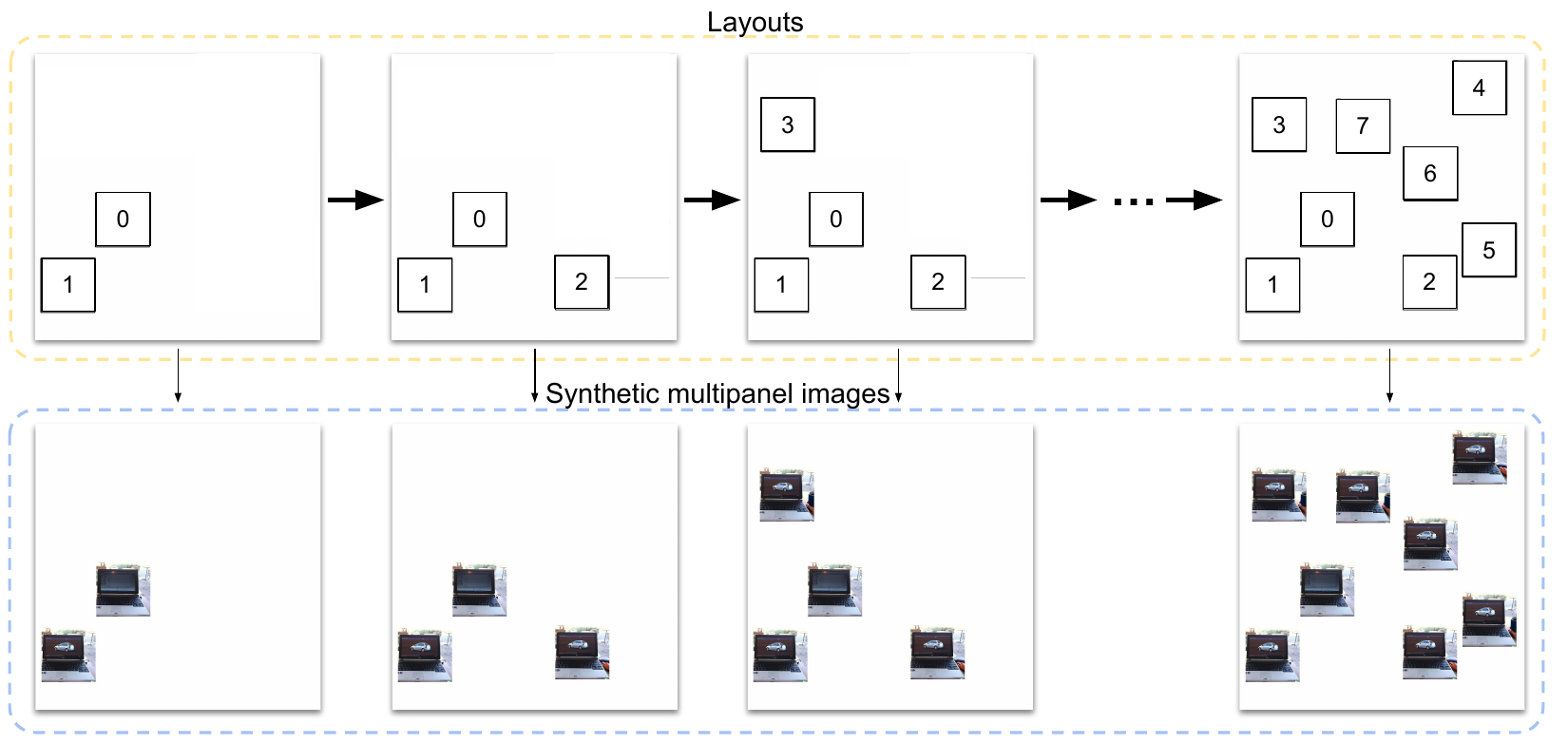}
    \caption{An example of the generation process for the layouts and synthetic multipanel images. When a new random subfigure position is determined, a new layout is formed. Based on the layouts, we position subfigures sequentially on a blank canvas according to a fixed order in each layout to create a synthetic multipanel image.}
    \label{fig:appe_fig_syn_process}
\end{figure*}

We illustrate this process in Figure~\ref{fig:appe_fig_syn_process}, where every time a new image is added to the blank canvas, a new synthetic multipanel image is created.

\begin{figure}[h]
\centering
\begin{minipage}[t]{0.5\textwidth}
\subfloat[]{
\includegraphics[width = \columnwidth]{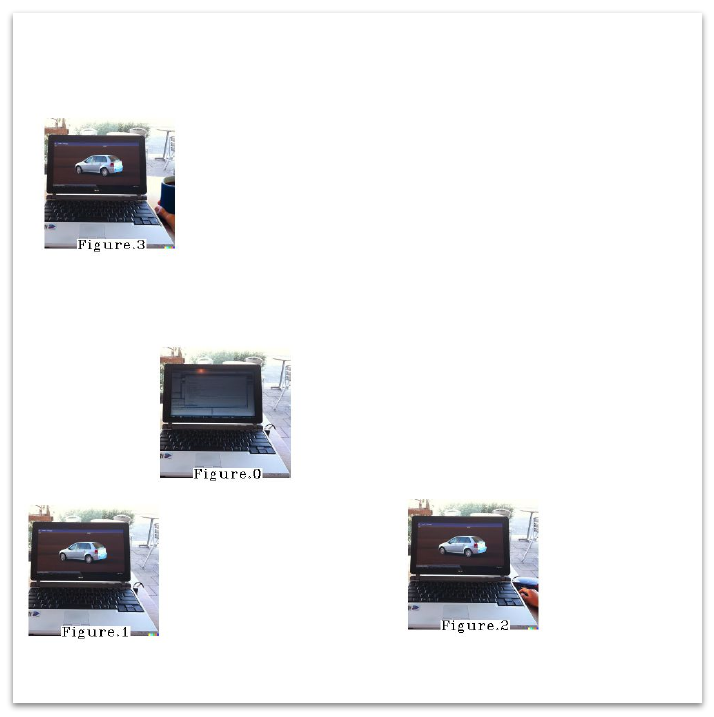}
\label{som_figure}}
\end{minipage}
\hspace{0.2cm}
\begin{minipage}[t]{0.5\textwidth}
\subfloat[]{
\includegraphics[width = \columnwidth]{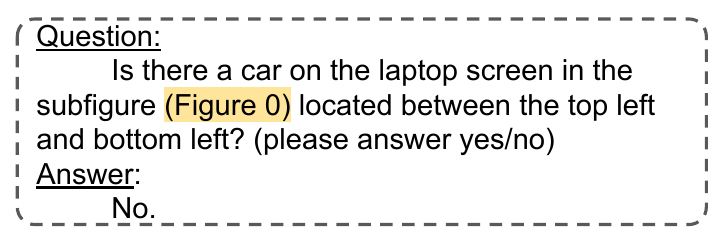}
\label{som_text}}
\end{minipage}

\caption{Example for (i) a multipanel image with subfigure captions including sequential numbers and (ii) a question and answer where the question explicitly refers the subfigure caption (highlighted ``Figure 0"). 
}
\end{figure}

\begin{figure}[t]
    \centering
    \includegraphics[width=\columnwidth]{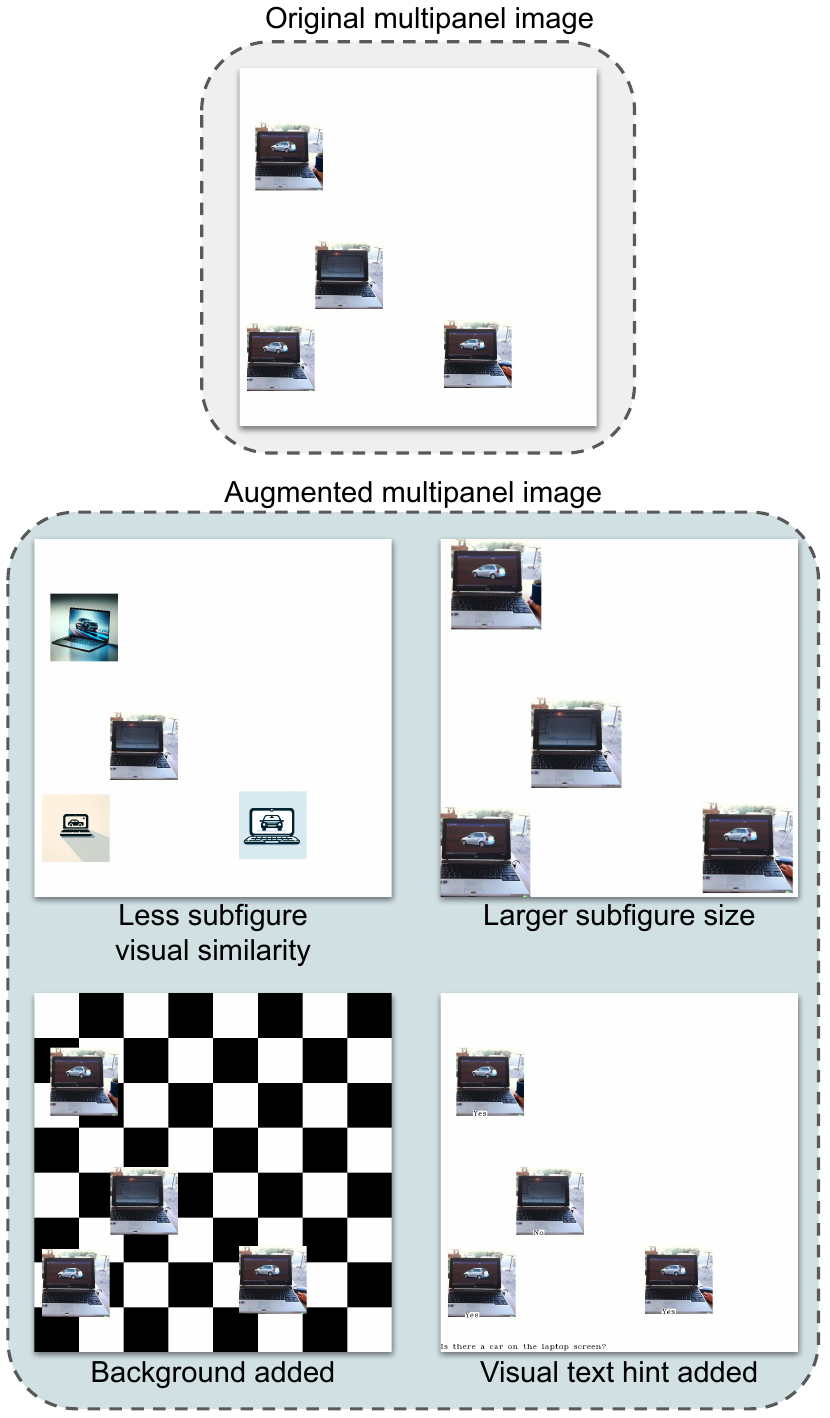}
    \caption{Examples of augmentations to synthetic multipanel images.}
    \label{fig:appen_aug}
\end{figure}

\begin{figure*}[th]
\centering
\setlength{\abovecaptionskip}{0.1cm}
\begin{minipage}[t]{\textwidth}
\subfloat[]{
\includegraphics[width=\textwidth]{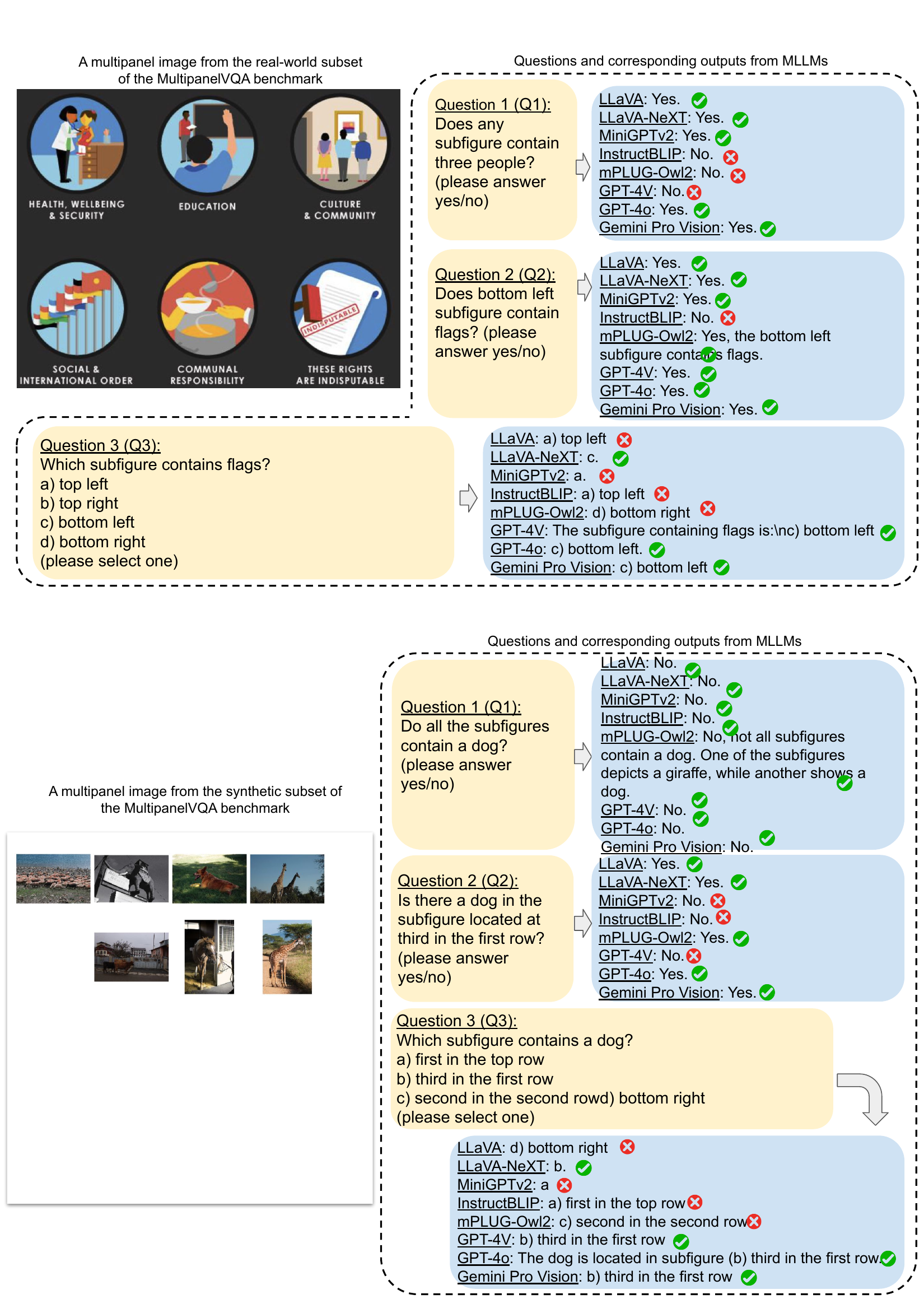}}
\end{minipage}

\begin{minipage}[t]{\textwidth}
\subfloat[]{
\includegraphics[width=\textwidth]{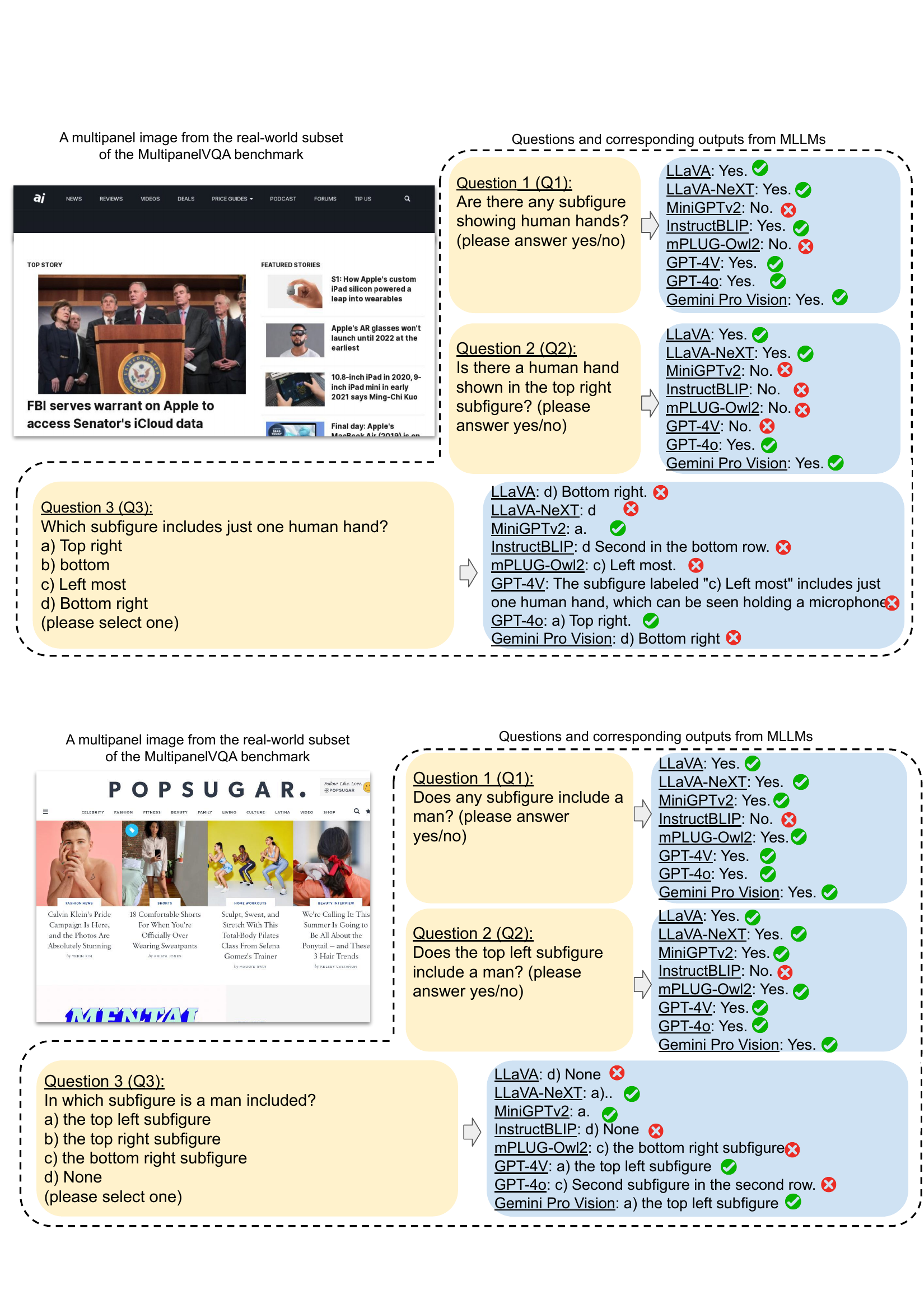}}
\end{minipage}

\caption{Samples of real-world multipanel images in the MultipanelVQA benchmark and outputs from models. (i) shows a poster multipanel image and (ii) shows a multipanel image of a web screenshot.
}
\label{examples}
\end{figure*}

\begin{figure*}[th]
\centering
\setlength{\abovecaptionskip}{0.1cm}
\begin{minipage}[t]{\textwidth}
\subfloat[]{
\includegraphics[width=\textwidth]{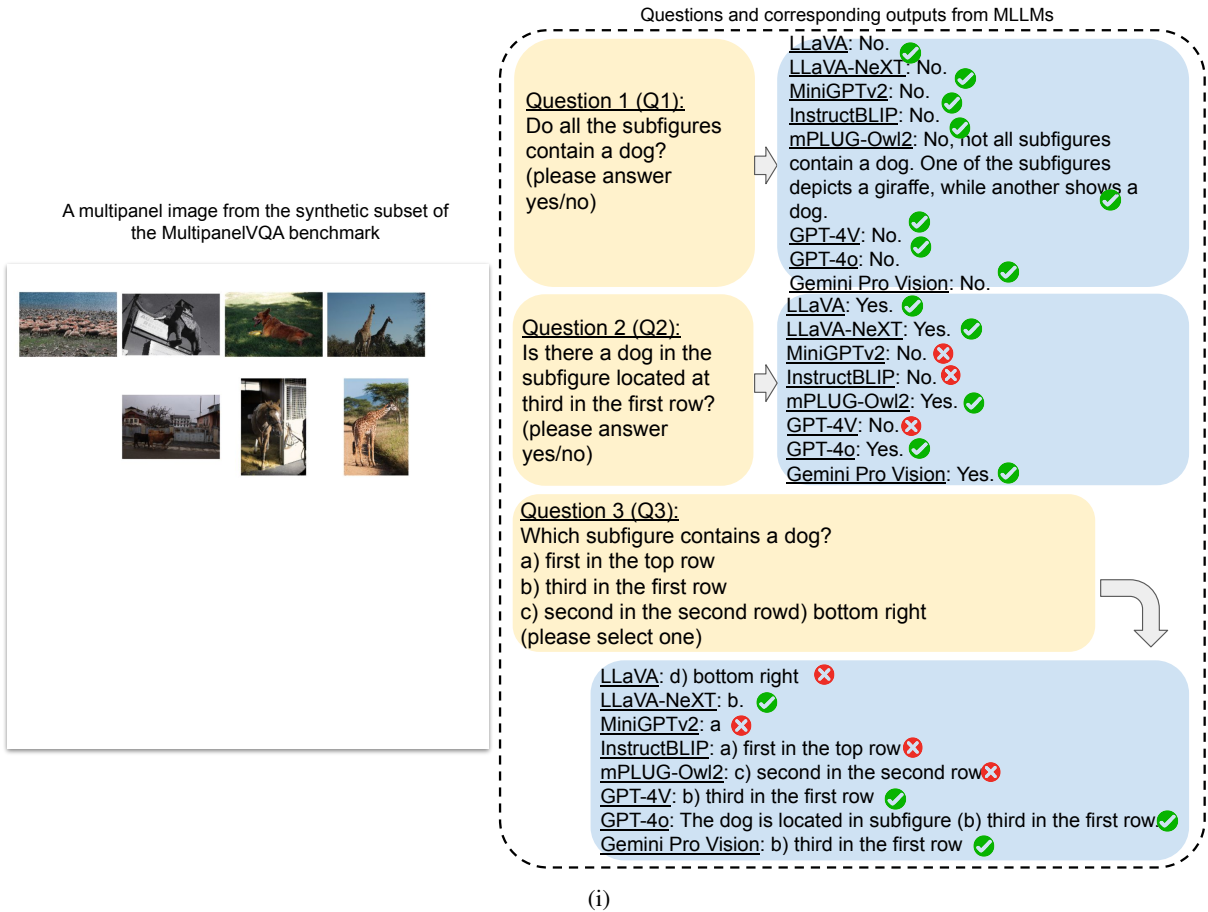}}
\end{minipage}

\caption{Sample of synthetic multipanel images in the MultipanelVQA benchmark and outputs from models.
}
\label{examples_synthetic}
\end{figure*}

\subsection{Question-Answer Generation}
\label{appen_qa_gen}
We prompt GPT-4 to generate three questions in three distinct styles and corresponding answers for each multipanel image, given the fact that all subfigures in a synthetic multipanel image come from the same image set in the source dataset and share a common question. The first question asks if all or any subfigure have a specific object or object attribute which is mentioned in the common question of the image set. The second and third will focus on the content of a specific subfigure, which is the one with a unique answer to the common question shared in the image set. 
The prompt, shown in Table~\ref{tab:appendix_prompt_llm} includes detailed instructions for how to generate the question-answer pairs while requiring information about the multipanel image which consists the subfigure numbers, the common question for the subfigures, the answer of the target subfigure to the common question and the positional description of the target subfigure which we manually annotate the positional description for each subfigure in advance. 

\subsection{Augmentation of the Synthetic Data Subset}
\label{appen_aug}
We augment the synthetic data subset of the MultipanelVQA benchmark to enable a more comprehensive evaluation of MLLMs performance on multipanel image understanding. The augmentation is done by involving new multipanel images that are altered from the original version in four different ways while keeping the corresponding questions and answers the same. First, we reduce the visual similarity among subfigures in multipanel images by generating new subfigures to replace the original ones. Since the original subfigures in each multipanel image come from the same image set of the source dataset, they share a visual similarity as they have a common question, and many even have the same answer to the common question. In order to reduce this similarity while keeping the questions and answers for the multi-panel image unaffected, we prompt DALL·E 3 \cite{betker2023improving} to generate various images that do not incur the same answer to the common question as the target subfigure and then replace the subfigures except the target subfigure with these newly generated images. As shown in Figure~\ref{fig:appen_aug}, in this way, subfigures in multipanel images, especially those based on MagicBrush \cite{Zhang2023MagicBrush} dataset, become less similar to each other visually. Second, we increase the subfigure size within the multipanel images by first removing some edge space for the multipanel image while keeping the ratio of height and width and then resizing the image to the original size. Third, we add a background with black and white chessboard patterns to every synthetic multipanel image, introducing a more complex visual backdrop. Last, we embed texts to the multipanel image, where these texts include the common question and the corresponding answers of each subfigure. 

\section{Samples of Model Outputs on Real-world Multipanel Images}
\label{more_examples}
We show some more real-world multipanel images of web screenshots and posters along with model outputs in Figure~\ref{examples}. Additionally, there an sample from the synthetic data subset in Figure~\ref{examples_synthetic}.

\section{Supported Input Image Resolutions of Tested MLLMs}
\label{image_res}
We show the supported input image resolutions of four tested open-sourced MLLMs in Figure \ref{fig:supported_res}. 
As illustrated in Figure \ref{fig:horizeontal}, the variation of input image resolution is a valid factor in model performance.

\begin{table}[t]
\centering
\resizebox{0.5\textwidth}{!}{
\begin{tabular}{lccc}
\hline
Models  & Input image resolution & \#visual tokens per input image \\
\hline
LLaVA 
& 336 & 576\\
LLaVA-NeXT 
& 672 & 576\\

MiniGPT-v2 
& 448 & 256\\
InstructBLIP 
& 224 & 256 \\
mPLUG-Owl 
& 224 & 256 \\

\hline
\end{tabular}}
\caption{Supported input image resolutions of tested MLLMs.}

\label{fig:supported_res}
\end{table}

\section{GPT-4 as Evaluator}
\label{GPT_evalu}
Given the output of MLLMs with the question and multipanel image as input, we prompt GPT-4 to judge if the output is a correct answer. The prompt is shown in Table~\ref{tab:appendix_evalu_prompt}, where the question, model's output and corresponding ground truth are inserted. If GPT-4's output is yes, we regard the model's output as correct and vice versa. 

\section{Examples of Subfigure Captions with Sequential Numbers as Visual Prompts}
\label{appen_som}
We experiment with adding captions to subfigures in the synthetic data subset of MultipanelVQA as a visual prompting method similar to the Set of Mark (SoM) visual prompting method \cite{yang2023setofmark}. The caption we add to the subfigures includes sequential numbers, as shown in Figure~\ref{som_figure}. Besides changing the multipanel images with subfigure captions, we also modify the corresponding questions to refer to the subfigure caption explicitly, as shown in Figure~\ref{som_text}.

\begin{table}[t]
    \centering

\begin{minipage}[t]{0.5\textwidth}
\subfloat[]{

    \resizebox{\linewidth}{!}{
    \begin{tabular}{llll}
    \toprule
        \multicolumn{4}{m{11cm}}{
          $\begin{array}{rl}
              \textbf{Prompt:} & \text{For question: }\{question\}\\
& \text{Compare the following answers:} \\
& \text{Text 1: } \{output\}\\
& \text{Text 2: } \{gt\}\\
& \text{Does the first one contain all key information in the second}\\
& \text{one? (yes/no)}\\
& \text{Answer:} \\

          \end{array}$
       }\\
         \bottomrule
    \end{tabular}}

}
\end{minipage}
\begin{minipage}[t]{0.5\textwidth}
\subfloat[]{

    \resizebox{\linewidth}{!}{
    \begin{tabular}{llll}
    \toprule
        \multicolumn{4}{m{11cm}}{
          $\begin{array}{rl}
              \textbf{Prompt:} & \text{\space \space \space \space For question: }\{question\}\\
& \text{\space \space \space \space Ground truth: } \{gt\}\\
& \text{\space \space \space \space Model predicted answer: } \{output\}\\
& \text{\space \space \space \space Based on the question and the ground truth answer, is the }\\
& \text{model's predicted answer correct? If multi-choice is provided, }\\
& \text{think about which choice is selected by the model, is it }\\
& \text{correct? (please answer yes/no)}
          \end{array}$
       }\\
         \bottomrule
    \end{tabular}}
}
\end{minipage}

    \caption{Text prompt for GPT-4 as an evaluator to judge if the output from the model $\{output\}$ is correct given the question $\{question\}$ ground truth answer $\{gt\}$. (a) shows the prompt for GPT-4 to evaluate the model output for the first and second types of question (Q1 and Q2) in MultipanelVQA. (b) shows the prompt for GPT-4 to judge the third type of question (Q3) in MultipanelVQA }

    \label{tab:appendix_evalu_prompt}
\end{table}

\begin{table*}[t]
    \centering
    \resizebox{\textwidth}{!}{
    \begin{tabular}{llll}
    \toprule
        \multicolumn{4}{m{25cm}}{
          $\begin{array}{rl}
              \textbf{Prompt:} & \text{\space \space \space \space You are asking questions about an multi-panel image composition with multiple subfigures. You will be given a description of the} \\
              & \text{overall layouts of the subfigures, a common question and answers to this question for each subfigure.} \\
            
& \text{\space \space \space \space First ask three questions (Q1, Q2, Q3) and then generate ground truth answers (A1, A2, A3) to each question.} \\

& \text{\space \space \space \space The second question (Q2) should be the same as the common question provided but specifically targeting at one subfigure. Make sure to include}\\
& \text{specific position of the subfigure targeted. } \\
& \text{\space \space \space \space The first question (Q1) asks if all or any subfigures have the specific object/attribute mentioned in Q2. (e.g. Do all the subfigures}\\
& \text{share certain object? Is there any subfigure that has a certain object?). }\\
& \text{\space \space \space \space For both answers A1 and A2, try not to refer to specific positions of subfigures and be concise.}\\

& \text{\space \space \space \space For the third question (Q3) make it a multi-choice question with a single answer based on the common question and answer. The answer (A3)}\\
& \text{should only be the subfigure targeted.} \\ 
& \text{\space \space \space \space Also generate a,b,c,d four choices and randomly put the correct answer in one of them, and fill the other choices with x. }\\
& \text{\space \space \space \space For the third answer (A3), only put in the label for the correct choice (a,b,c or d).
Ask questions only based on the direct information you}\\
& \text{get from the provided common question and answers.}\\
& \text{\space \space \space \space At the end of each question (Q1, Q1 or Q3), indicate what kind of answer is needed for the question. (eg. please answer yes/no, please select one).}\\
& \text{\space \space \space \space Answers generated should be consice without any explanation.}\\
& \text{\space \space \space \space Your output should be in the following format: Q1: A1: 
Q2: 
A2: 
Q3: 
A3: } \\
& \text{\space \space \space \space There are } {\{num\_subfigure\}}\text{ subfigures in the image. The common question for all subfigures are: }\{com\_question\} \text{.}\\
& \text{\space \space \space \space The answer from the target subfigure is: }\{answer\_target\_subfigure\} \text{.} \\
& \text{\space \space \space \space The answer for the other subfigures are not the same as the target subfigure. Ask questions about the target subfigure located at }\{pos\_description\} \text{.}\\
              
          \end{array}$
       }\\
         \bottomrule
    \end{tabular}}
    \caption{Text prompts for generating questions and answers of multipanel images in the synthetic subset of MultipanelVQA benchmark. $\{num\_subfigure\}$ is the number of subfigures in the multipanel image. $\{com\_question\}$ is the common question in the image set from source datasets. $\{answer\_target\_subfigure\}$ is the answer of the target subfigure to the common question, which is different from the answer from the other subfigures selected. $\{pos\_description\}$ is the position description for the target subfigure predefined by human. }
    \label{tab:appendix_prompt_llm}
\end{table*}

\end{document}